\documentclass[letterpaper, 10 pt, conference]{ieeeconf}  

\IEEEoverridecommandlockouts                              

\usepackage{color,array}

\usepackage{graphicx}
\usepackage{booktabs}

\usepackage{newfloat}
\usepackage{listings}
\usepackage{multirow}
\usepackage{amsmath,amssymb,amsfonts}
\usepackage{diagbox}
\usepackage{booktabs}
\usepackage{subcaption}

\title{\LARGE \bf
Federated Joint Learning of Robot Networks in Stroke Rehabilitation
}

\author{Xinyu Jiang$^{1}$, Yibei Guo$^{1}$, Mengsha Hu$^{2}$, Ruoming Jin$^{2}$, Hai Phan$^{3}$, Jay Alberts$^{4}$, Rui Liu$^{1*}$
\thanks{$^{1}$ are with the College of Aeronautics and Engineering, Kent State University, Kent, Ohio 44242, USA. $^{2}$ is with the Department of Computer Science, Kent State University, Kent, Ohio 44242, USA. $^{3}$ is with the New Jersey Institute of Technology, Department of Data Science, Newark, New Jersey 07102, USA. $^{4}$ is with the Cleveland Clinic, Concussion Center, 9500 Euclid Ave, Cleveland, OH 44195, USA. $^{*}$Rui Liu is the corresponding author, email: ruiliu.robotics@gmail.com.}
}

\begin{document}

\maketitle
\thispagestyle{empty}
\pagestyle{empty}

\begin{abstract}

Advanced by rich perception and precise execution, robots possess immense potential to provide professional and customized rehabilitation exercises for patients with mobility impairments caused by strokes. Autonomous robotic rehabilitation significantly reduces human workloads in the long and tedious rehabilitation process. However, training a rehabilitation robot is challenging due to the data scarcity issue. This challenge arises from privacy concerns (e.g., the risk of leaking private disease and identity information of patients) during clinical data access and usage. Data from various patients and hospitals cannot be shared for adequate robot training, further compromising rehabilitation safety and limiting implementation scopes. To address this challenge, this work developed a novel federated joint learning (FJL) method to jointly train robots across hospitals. FJL also adopted a long short-term memory network (LSTM)-Transformer learning mechanism to effectively explore the complex tempo-spatial relations among patient mobility conditions and robotic rehabilitation motions. To validate FJL's effectiveness in training a robot network, a clinic-simulation combined experiment was designed. Real rehabilitation exercise data from 200 patients with stroke diseases (upper limb hemiplegia, Parkinson's syndrome, and back pain syndrome) were adopted. Inversely driven by clinical data, 300,000 robotic rehabilitation guidances were simulated. FJL proved to be effective in joint rehabilitation learning, performing 20\% - 30\% better than baseline methods.

\end{abstract}

\section{INTRODUCTION}

Stroke is a global healthcare problem contributing to individual disability and death \cite{dobkin2004strategies}; rehabilitation typically aims to train patients in compensatory strategies with proximal (e.g., shoulder abduction, arm flexion) and distal (e.g., hand open, finger extension) movements to facilitate patient recovery on strength, speed, endurance, and precision of multijoint movements \cite{dobkin2004strategies, langhorne2011stroke}. While training a human expert for professional rehabilitation is expensive and lengthy, for example, an attending physician-level expert will need an average professional education and training time of 8-11 years and 0.2-0.5 million dollars \cite{carpino2018assessing}. 

Powered by sensor and control technologies, robots can precisely and durably provide patient training exercises to ensure quality rehabilitation exercise, significantly reducing human workload and economic/time costs; most importantly, powered by the latest learning algorithms, a robot with expert-level skills can be trained within several days \cite{karoly2020deep}. Therefore, it is promising to rely on robots for durable, reliable, and economical rehabilitation for movement disorder stroke diseases, such as hemorrhagic stroke and hemiplegic stroke \cite{chang2013robot}. 

\begin{figure}[!t]
	\centering{
	\includegraphics[width=.8\linewidth]{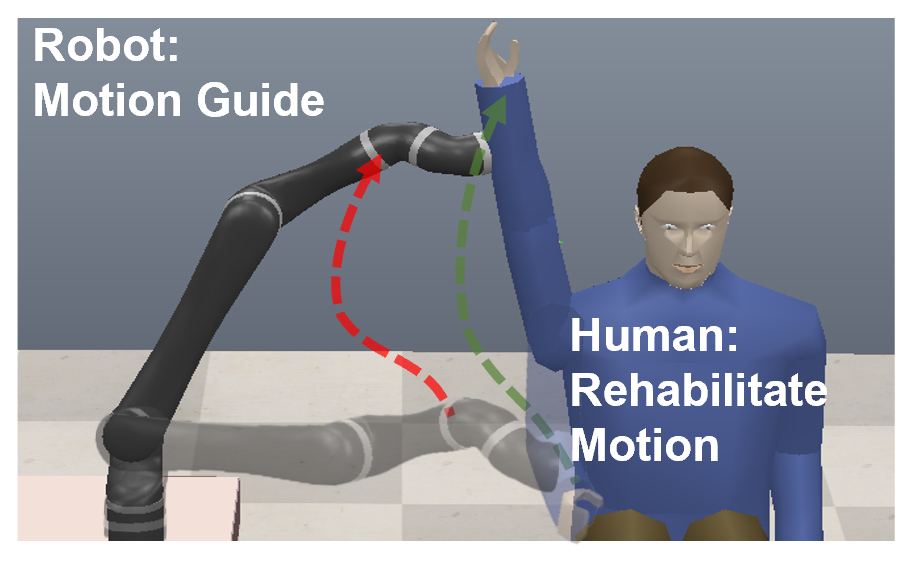}
        \setlength{\belowcaptionskip}{-10pt}
        \setlength{\abovecaptionskip}{0pt}
	\caption{Settings for robot-guided rehabilitation.}
        \vspace{-20pt}
	\label{settings_a}
	}
\end{figure}

However, training a professional and safe rehabilitation robot is challenging due to clinical data scarcity. Besides treatment-relevant information (e.g., stroke types and motor impairments), clinical rehabilitation data also includes irrelevant but private information (e.g., patients' identity, physiological characteristics, and other illnesses) \cite{veliz2020not}. Restrained by concerns of leaking patient information, clinical data cannot be accessed across hospitals; small-amount local data inadequately train a rehabilitation robot, further undermining its performance and safety and impeding widespread implementations of robotic rehabilitation \cite{al2023review}. Besides, patients vary in physical characteristics and motor impairments, adding challenges for robots to provide customized rehabilitation \cite{hussain2020state}.

Therefore, to address the data scarcity issue, in this research, a novel joint training method -- \textit{Federated Joint Learning (FJL)} was developed to collaboratively train robots crossing hospitals. Particularly, our work in this paper mainly has three contributions:

\begin{itemize}
    \item A federated joint learning network was developed to network robots crossing hospitals and enable them to mutually learn rehabilitation skills from each other without directly accessing original patient data. Fig. \ref{settings_a} illustrates the simulation environment settings.

    \item A LSTM-Transformer learning framework was developed to efficiently extract representative motion plans from complex spatiotemporal motions of patient joints with differences in body characteristics and motor impairment degree.

    \item A novel relational loss was designed to refine the robot pose estimation result and improve the accuracy of the pose estimation model.
\end{itemize}

\section{Related Works}

\subsection{Robotic Rehabilitation: Benefits and Challenges}

Recently, robotic rehabilitation has presented an increasing trend for enhanced patient care, particularly benefiting those requiring long-term treatments like stroke (Upper limb hemiplegia). In \cite{deneve2008control}, a rehabilitation and training robot for upper limbs allows the execution of the sequence of switching control laws corresponding to the training configuration, which demonstrates the effectiveness of robotic rehabilitation. Meanwhile, in medical rehabilitation, the inherent complexities present notable challenges. First, rehabilitation demands an intricate understanding of temporal action spaces and variable patient conditions \cite{pierella2020multimodal}. Second, crafting an appropriate treatment becomes challenging as it needs to be customized to an individual with unique psychological requirements, which the above methods neglect. \cite{johnson2007potential} built systems with personalized rehabilitation support for post-stroke patients but still require human efforts to complete the rehabilitation. Notably, these methods mentioned have revealed a significant cost overhead and remain labor-intensive \cite{lo2019economic}.

\subsection{Clinical Data Accessibility Difficulty}

Accessing clinical data for robot/AI training is challenging due to privacy concerns during data usage. The clinical data collected by various sensors (e.g., acceleration, motion, camera, and voice) describe both the stroke information and other private information of patients, such as identity, physiological characteristics (e.g., height, arm length), and likely other symptoms and illness \cite{boone2011judicial, henriksen2013privacy}. This concern over leaking patient information undermines the feasibility of sharing clinical data for robot/AI training; while exploring large-scale clinic data is highly desired for comprehensively exploring treatment plans behind various patient conditions\cite{price2019privacy}\cite{wang2022efficient}\cite{sun2017revisiting}. To address this issue, \cite{sweeney2002k} transformed raw data into simplified formats by removing information details and granularity. Differential privacy (DP) \cite{abadi2016deep}, which describes patterns and trends instead of details about specific individuals, was used to add noise to the data without losing statistical information. However, those approaches undermine data integrity and require careful model tuning to reach desired performance. For medical rehabilitation, a comprehensive exploration of spatial-temporal motion dependencies is needed to ensure rehabilitation safety; information missing could potentially bring safety issues into motion planning (e.g., aggressive actions, sudden motion transitions). In this research, FJL will enable comprehensive rehabilitation learning without compromising data integrity or exposing patient conditions. 

\subsection{Clinical Data Accessibility Difficulty}
Accessing clinical data for robot/AI training is challenging due to privacy concerns during data usage. The clinical data collected by various sensors (e.g., acceleration, motion, camera, and voice) describe both the stroke information and other private information of patients, such as identity, physiological characteristics (e.g., height, arm length), and likely other symptoms and illness \cite{boone2011judicial, henriksen2013privacy}. This concern over leaking patient information undermines the feasibility of sharing clinical data for robot/AI training; while exploring large-scale clinic data is highly desired for comprehensively exploring treatment plans behind various patient conditions\cite{price2019privacy}\cite{wang2022efficient}\cite{sun2017revisiting}. To address this issue, \cite{sweeney2002k} transformed raw data into simplified formats by removing information details and granularity. Differential privacy (DP) \cite{abadi2016deep}, which describes patterns and trends instead of details about specific individuals, was used to add noise to the data without losing statistical information. However, those approaches undermine data integrity and require careful model tuning to reach desired performance. For medical rehabilitation, a comprehensive exploration of spatial-temporal motion dependencies is needed to ensure rehabilitation safety; information missing could potentially bring safety issues into motion planning (e.g., aggressive actions, sudden motion transitions). In this research, FJL will enable comprehensive rehabilitation learning without compromising data integrity or exposing patient conditions. 

\subsection{Federated Robot Learning}
Federated learning is an emerging method for secure and decentralized robot training. Instead of centralizing data for training as in traditional paradigms, federated learning allows robots to train on local data, sharing only model updates, thus preserving data privacy and cutting transmission costs \cite{li2021survey}.
This emphasis is also evident in healthcare applications, where federated learning has been employed to preserve patient information. For instance, \cite{dou2021federated} employed it for detecting COVID-19-related CT abnormalities, and \cite{chen2020fl} for privacy-focused drug discovery.
From the perspective of robotic learning, federated learning is beneficial when robots operate in diverse environments but need to collaborate for better performance. Beyond ensuring data security, federated learning allows users to decide on sharing their model updates, recognizing the advantages of collaborative training.
Several applications in robotics have leveraged federated learning, including multi-robot cooperative tasks \cite{fan2021fault}, enhancing the efficiency of autonomous driving tasks \cite{liang2022federated}, and human-robot interaction scenarios \cite{su2022deploying}.
In robotic rehabilitation, federated learning is motivated by two main factors. First, patient data vary widely across different hospitals or facilities. Second, there is a need for personalized therapy that still benefits from collective insights. By integrating federated learning, the goal is to empower robot training across multiple hospitals to learn and enhance the rehabilitation capability collaboratively. Additionally, the approach aims to ensure that robots do not directly access patient data to preserve its integrity. Therefore, it lays a stronger foundation for the future cross-hospital deployment of AI-powered medical robots.

\section{Our Approach: FJL Network}
\subsection{Problem Statement}
Constrained by patient-sensitive data concerns, rehabilitation data from different hospitals cannot be shared to train robots, limiting robotic rehabilitation training to a small data scale with unreliable performance. To address this data scarcity problem, a novel joint training method – Federated Joint Learning (FJL), is developed to train robots across hospitals collaboratively. FJL tackles three pivotal questions in robotic rehabilitation learning: \textit{1). How to achieve joint learning for training multiple robots without accessing the patient's sensitive data. 2). How to estimate robot joint pose estimation accurately.  3). If the output pose estimation result is not accurate, how to refine the joint pose estimation result.}

\textit{Federated Joint Learning Problem.} 
The objective of federated joint learning is training robots with mutual skill exchange without directly accessing patient data. In the FJL architecture, the original patient data is invisible to external robots, and robot learners (servers) update model parameters for rehabilitation action learning. To be concrete, the problem can be explained as follows - a dedicated learner node (a robot) called server (S) devises the objective of a globally learned model (GLM) to be built. It decides the model's architecture denoted by  ${f_\theta }:X \to Y$. The tuple 
$\left( {x,y} \right) \in \left( {X,Y} \right)$ denotes a sample input-output pair. A loss function models the objective of the server
$\mathcal{G}\left( \theta \right)$. In this paper, multiple robots use its local private dataset $R_1$, ... $R_n$ to train the customized models $\mathcal{M}_1$,....$\mathcal{M}_n$ such that the prediction result $\hat{y}_1$,  ..., $\hat{y}_m$ are closed to the ground truth $y_1$, ...$y_n$, respectively; not any patient personal data is exposed to other clients or any other users directly. 

\textit{Joint Pose Estimation Problem.} The objective is to estimate the robot's joint pose and generate a robot rehabilitation plan according to the patient's joint pose input. In this FJL architecture, a robot should generate accurate pose estimation to help patient rehabilitation accurately. The error should be smaller than 0.1\textit{m}. The patient's joint pose in rehabilitation exercises is described by 7-dimensional vector joint position $x$, $y$, $z$, and orientation $q_x$, $q_y$, $q_z$, $q_w$. denoted as \textit{input} $X = \left\{ {x_t^i} \right\} \in {\mathbb{R}^{7 \times T}}$ as N related univariate time series joint pose, where $T$ is the number of timestamps, and $Y = \left\{ {y_t^i} \right\} \in {\mathbb{R}^{6 \times 1}}$ is the output estimation result. The robot joint pose estimation problem can be described as: learning a mapping function $\mathcal{f}$ that maps the historic $P$ timesteps patients' joint pose input $\left[ {{X_{t - P + 1}},{X_{t - P}},...,{X_t}} \right]$ into the future robot estimation of next steps, as shown in Eq.(\ref{spatial_temproal_problem}). 
The output $Y_t$ includes the robot joint position $x_r$, $y_r$, $z_r$ and velocity $v_{x_r}$, $v_{y_r}$, $v_{z_r}$.

\begin{equation}
\label{spatial_temproal_problem}
\left[ {{X_{t - P + 1:t}}} \right]\xrightarrow{f}\left[ {{Y_t}} \right]
\end{equation}

\begin{figure}[!tbp]
	\centering{
	\includegraphics[width=1.0\linewidth]{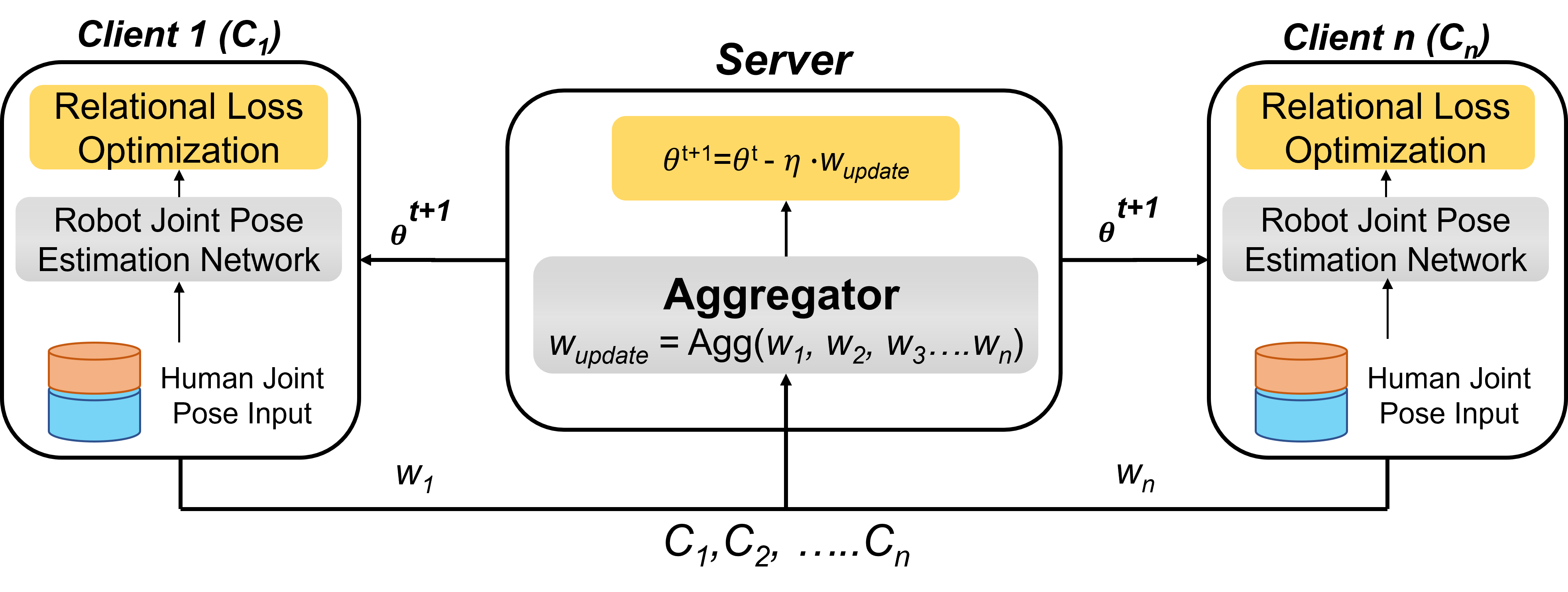}
        \setlength{\abovecaptionskip}{-10pt}
        \setlength{\belowcaptionskip}{-10pt}
	\caption{Federated joint learning framework overview.}
	\label{frameworkoverview}
	}
\end{figure}

\subsection{Solution: FJL Network}

In the FJL framework, there are mainly three modules - 1) A Federated Joint Learning module to realize the joint learning to train rehabilitation robots jointly and concealing patients' sensitive data while sharing the learning data. 2) A Robot Joint Pose Estimation module provides accurate robot pose estimation according to the input patients' joint pose based on the LSTM-Transformer module. 3) An relational-based robot estimation refinement module to optimize the robot pose estimation network parameters $\theta$ that is described in 2). For a single robot, the input data is the patient's joint pose, and the output data is the robot's joint pose. The data uploaded to the server is updated gradient through federated joint learning modules. The framework is shown in Figure \ref{frameworkoverview}.

\begin{figure}[!tbp]
	\centering{
	\includegraphics[width=0.88\linewidth]{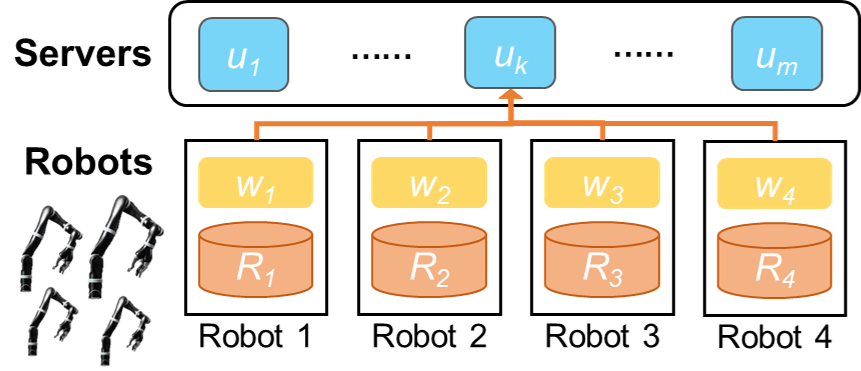}
        \setlength{\belowcaptionskip}{-10pt}
	\caption{The federated learning-based multi-robots joint learning framework.}
	\label{jointlearning}
	}
\end{figure}

\noindent\textbf{Robot Federated Joint Learning.} In the previous architecture of multi-robot training, the prerequisite for efficiently training multiple robots is to share the training data between them, along with a random shuffle. This approach enables the requirement of identically distributed data (i.i.d) that will reduce the model's performance. However, considering the patient's data leakage, it is infeasible to share data with robots directly. Even using state-of-the-art encryption algorithms (RSA encryption), the risk of leaking private information is still high, such as inferring patient identity and other illnesses.

The proposed joint learning architecture is shown in Figure \ref{jointlearning} to train the multi-rehabilitation robots and conceal patients' sensitive data simultaneously. To simultaneously train multiple robots, federated learning (FL) architecture will be used to jointly learn actions from human expert demonstrations while maintaining the original data invisible to robot learners. Figure \ref{jointlearning} shows the FL joint training framework. In this framework,  the objective task is to define an objective training function $F_i$ that maps the model parameters set ${\theta _i} \in {\mathbb{R}^d}$ to a training loss respect to private $i$-th robot data $R_i$.
\begin{equation}
\mathop {\min }\limits_\theta \left\{ {\mathcal{G}\left( \theta \right): = \sum\limits_{i = 1}^m {{F_i}\left( {{\theta_i}} \right) + \lambda \sum\limits_{i < j}^m {A\left( {\left\| {{\theta_i} - {\theta_j}} \right\|} \right)} } } \right\}
\end{equation}

\noindent where $A\left( {{{\left\| {{\theta _i} - {\theta _j}} \right\|}^2}} \right)$ is an attention including function that described in \cite{huang2021personalized} to measure the difference between $\theta_i$ and $\theta_j$ in a non-linear manner, and $\lambda$ is a normalized parameter. 

However, training robots locally and only uploading gradients will cause non-iid problems \cite{huang2021personalized} and degradation of the neural network, and reduce the neural network's performance. A similar architecture described in \cite{mcmahan2017communication} is used to overcome this disadvantage. For every iteration, the clients will update the training gradient $\delta_{C_i}$, an average aggregation strategy is used to compute the updated gradient:

\begin{equation}
\label{equation:federated_min}
\delta _{agg} = \frac{1}{n}\sum\limits_{i = 1}^n {{\delta _{{C_i}}}}
\end{equation}

The Eq.\ref{equation:federated_min} shows that all of the original sensitive data are transformed to a gradient update, which is updated to the server at last, and patients' sensitive data is invisible to all clients and servers. 

At last, the gradient that gets from clients is aggregated and updated by using the following formula:
\begin{equation}
{\theta ^{update}} = \theta  - \eta  \cdot {\delta _{agg}}
\end{equation}

\noindent\textbf{Robot Joint Pose Estimation.} LSTM is a classical method to deal with spatial-temporal data. However, the lack of seq2seq and attention mechanism makes it hard to find the latent relationship in such data. Multi-head attention is a good mechanism to mine the relationship in spatial-temporal data and improve the accuracy of the prediction result \cite{pierella2020multimodal}. So, a multi-head attention-based module - Transformer is introduced to Robot Joint Pose Estimation Network.

\begin{equation}
{\hat Y_t} = f\left( {{X_{t - P + 1:t}}} \right)
\end{equation}

\begin{figure}[!tbp]
	\centering{
	\includegraphics[width=1.0\linewidth]{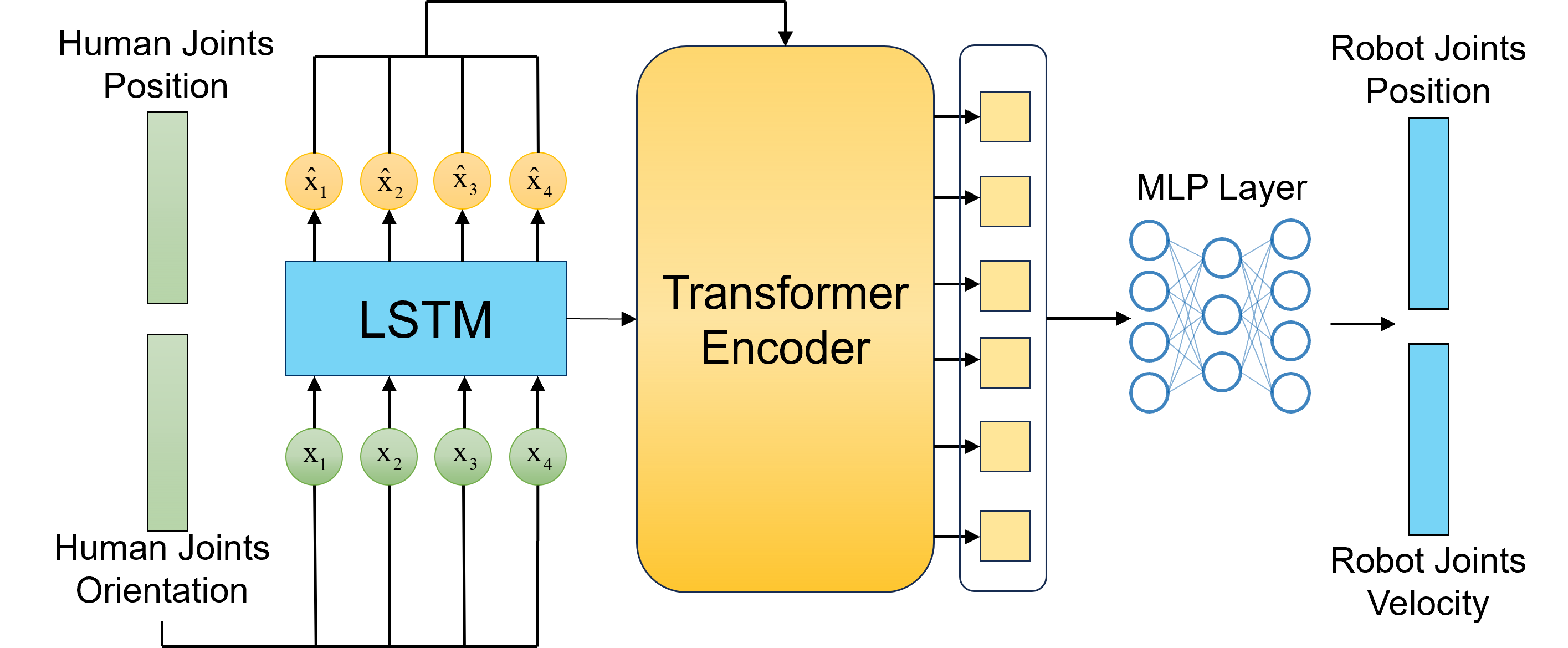}
        \setlength{\belowcaptionskip}{-10pt}
	\caption{Robot joint pose estimation network pipeline.}
	\label{fig:robot_pose_estimation}
	}
\end{figure}

\noindent where in the robotic arm settings, $X=\{J_f^n, J_p^n, J_i^n, J_v^n\}$ represents the state of robot with $n$ joints, including $J_f$ as joint's force, $J_p$ as joint's position in space, $J_i$ as joint's intrinsic position, and $J_v$ as joint's velocity in space. These states are controllable via step motors and thus enable the robot arm to perform the designated motions.

As shown in Figure \ref{fig:robot_pose_estimation}, the proposed LSTM-Transformer-based robot joint pose estimation module consists of three parts. The first layer is an LSTM model to encode patients' joint input and generate an $m \times n$ feature embedding. Then, based on the output features, a Transformer module is applied to predict the robot pose estimation result and improve the accuracy of the estimation result. At last, multiple MLP layers are used to get the robot joint estimation result. The model computation procedure $f\left( {{X_{t - P + 1:t}}} \right)$ can be described as:

\begin{equation}
{X_{concat}} = {\text{concat}}\left( {{X_{t - p + 1:t}}} \right)
\end{equation}

\begin{equation}
{X_{lstm}},{h_{lstm}} = {\text{LSTM}}\left( {{X_{concat}},h_{lstm}^{i - 1}} \right)
\end{equation}

\begin{equation}
{{\hat Y}_t} = {\text{TransformerBlock}}\left( {{X_{lstm}}} \right)
\end{equation}

At last, an MSE loss function is applied to train the robot joint estimation model:
\begin{equation}
{\mathcal{L}_{MSE}} = \frac{1}{N}\sum\limits_{i = 1}^N {{{\left( {f(X_{t - P + 1:t}) - {Y_t}} \right)}^2}} = \frac{1}{N}\sum\limits_{i = 1}^N {{{\left( {\hat Y_t - {Y_t}} \right)}^2}}
\label{equation:mse_loss}
\end{equation}

\noindent where $Y_t$ is the t-th robot joint ground truth and $N$ is the total$N$-timestamp.

\noindent\textbf{Relational Loss.} Due to the control error in the robot system, robot pose estimation from human rehabilitation requirements contains accurate errors and noise, undermining the accuracy of robot motions and counterproductive patient rehabilitation quality. According to work \cite{yang2023trajectory}, errors for rehabilitation motions need to be within 10 centimeters to ensure good recovery progress. To improve the accuracy of pose estimations, a refinement mechanism based on a relational surrogate loss \cite{huang2022relational} will be developed.

PCK value metrics is a metrics tool to measure the joint estimation error according to the output prediction accuracy. To accurately predict human pose, PCK value metrics will be used to measure the Euclidean distance of joint positions during movements and, therefore, accurately identify joints. The evaluation formula is Eq.(\ref{equation:pck_value}):

\begin{figure}[tbp]
    \centering 
\begin{subfigure}{0.23\textwidth}
  \includegraphics[width=\linewidth]{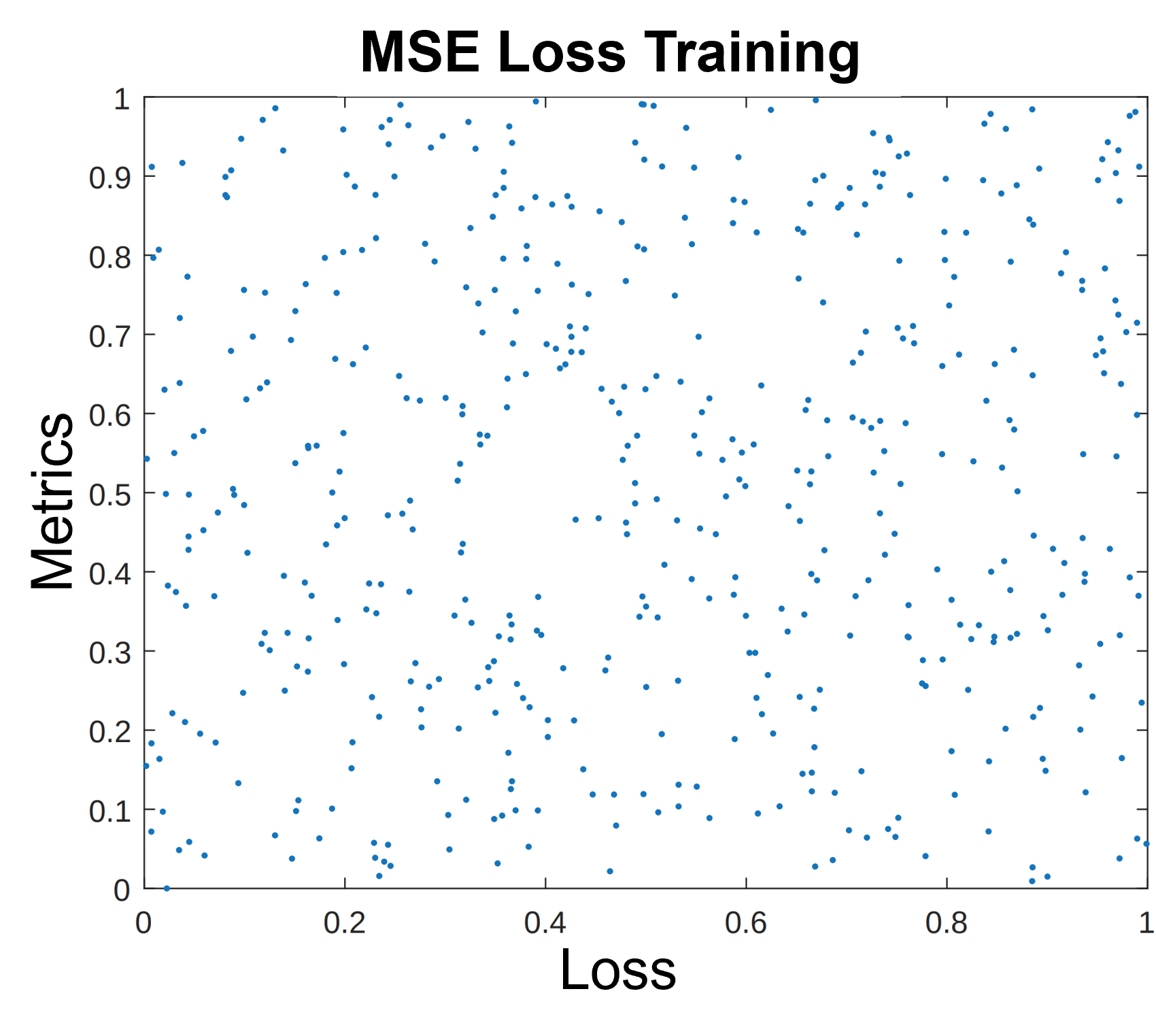}
  \setlength{\abovecaptionskip}{-10pt}
  \caption{MSE loss only.}
\end{subfigure}
\begin{subfigure}{0.23\textwidth}
  \includegraphics[width=\linewidth]{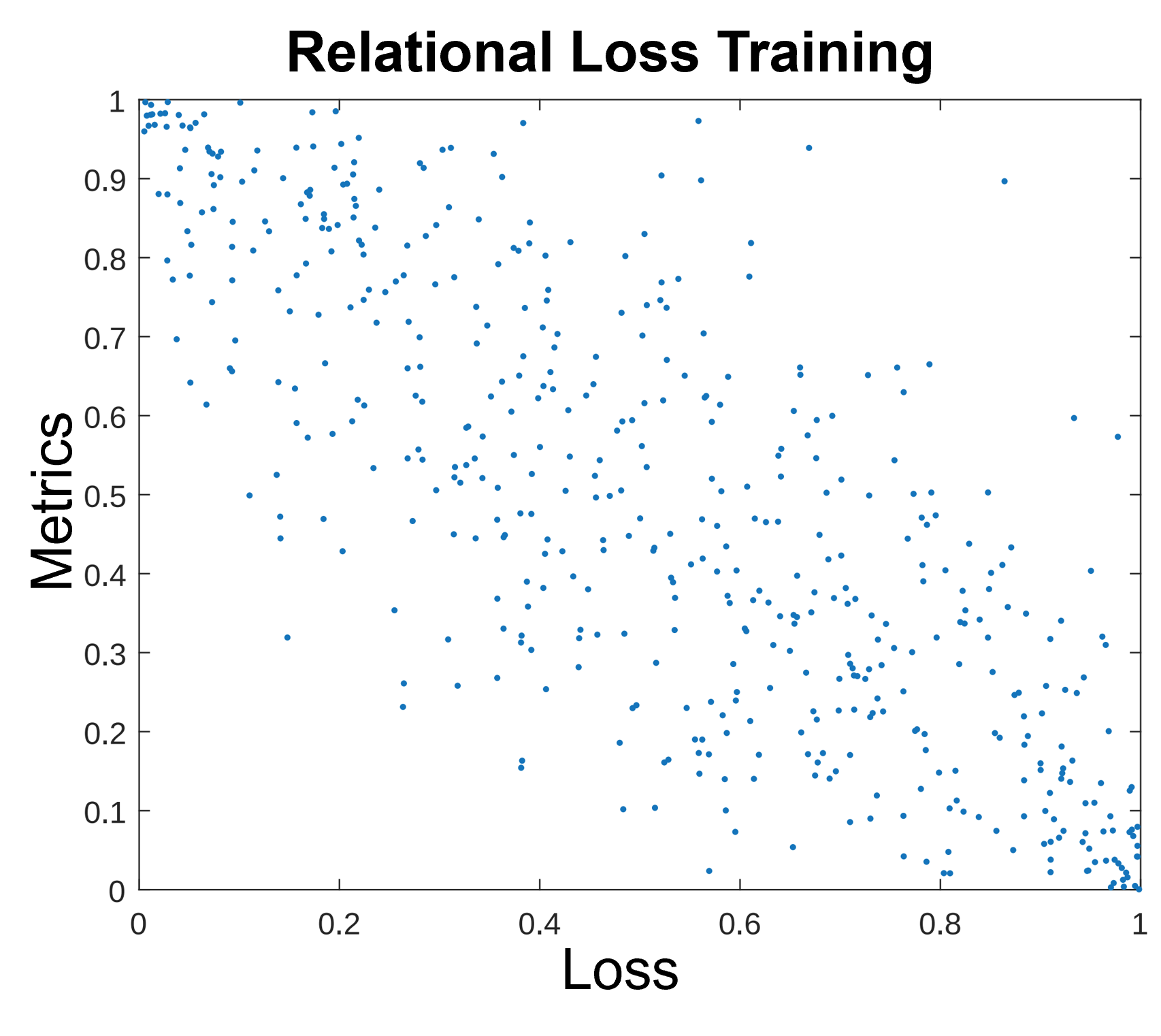}
  \setlength{\abovecaptionskip}{-10pt}
  \caption{Reloss refining.}
\end{subfigure}
\setlength{\abovecaptionskip}{5pt}
\caption{Loss and metrics tendency. After model refinement, the MSE loss and metrics become negative-correlated.}
\label{fig:negative_correlation_summary}
\vspace{-10px}
\end{figure}

\begin{equation}
\label{equation:pck_value}
PCK_i^k = \frac{{\sum\nolimits_p {\delta \left( {\frac{{{d_{pg}}}}{{d_p^{def}}} \le {T_k}} \right)} }}{{\sum\nolimits_p 1 }}
\end{equation}

\noindent where $d_{pg}$ is the Euclidean distance between the pose estimation result and ground truth according to the joint position $x_p$, $y_p$, $z_p$ and velocity $vx_p$, $vy_p$, $vz_p$.

To utilize PCK pose evaluation to update pose estimation network parameters $\theta$, MSE loss used in Eq. \ref{equation:mse_loss} will integrate a negative correlation evaluation value which name is Spearman's correlation and construct a relational loss to refine the joint pose estimation network parameters $\theta$. 

\begin{figure*}[!htb]
	\centering{
	\includegraphics[width=1.0\linewidth]{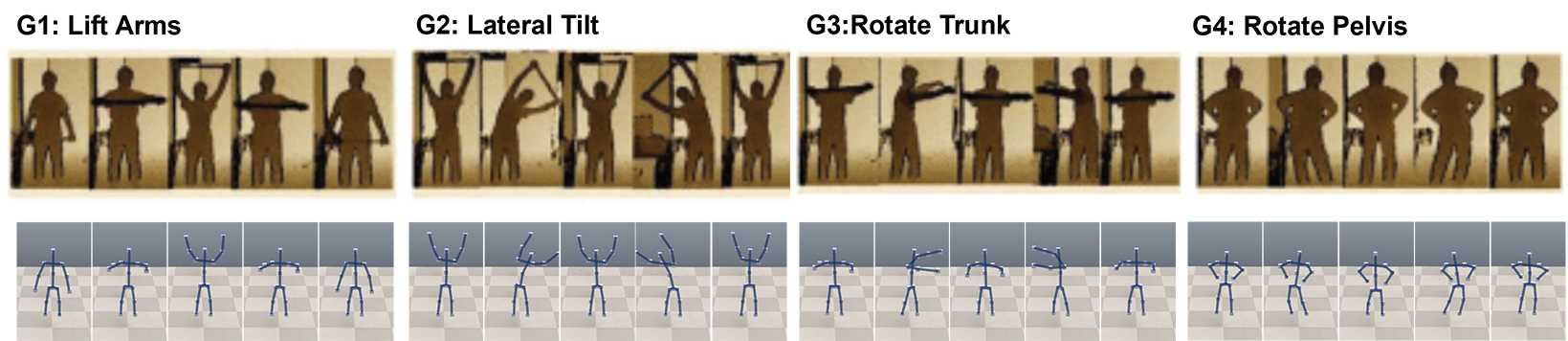}
        \setlength{\belowcaptionskip}{-10pt}
        \setlength{\abovecaptionskip}{0pt}
	\caption{Human rehabilitation exercise data from different patients and experts: clinical data (upper row) and simulated data (lower row). The data includes expert demonstration and patients with three different kinds of pain or postural disorder: Upper limb hemiplegia; Parkinson's syndrome; and back pain syndrome. In the experiments, data from 200 patients with four types of exercises were used. }
	\label{settings_joints}
	}
\end{figure*}

MES and PCK metrics are negatively corrected that smaller MSE loss of neural network will generate higher PCK metrics which means a more accuracy pose estimation.

In an ideal situation, there is a negative correlation between the MSE loss and PCK metrics - that is, during training, the smaller the MSE loss of the neural network, the higher the PCK metrics should be, as shown in Figure. \ref{fig:negative_correlation_summary}(b). However, in general, there is no strict negative correlation between MSE loss and metrics, as shown in Figure. \ref{fig:negative_correlation_summary}(a). This lack of correlation leads to inconsistencies in the network model, thereby providing us with an opportunity to combine metrics and MSE loss for network parameter optimization.

To evaluate the correlation between the MSE loss and PCK metrics, the most commonly used Spearman's rank correlation according to the distribution of two random samples. The Spearman's value smaller than 0 indicates two samples have a negative correlation; conversely, it indicates a positive correlation. To this end, to make MSE loss and PCK metric negative correlation, minimizing the Spearman's value is the final objective. For two vectors \textbf{$a$} and \textbf{$b$} with size $n$, the Spearman's correlation rank correlation is defined as: 
\begin{equation}
\label{equation:spearman}
\rho \left( {a,b} \right) = \frac{{\frac{1}{n}\sum\nolimits_{i = 1}^n {\left( {{r_{ai}} - E\left( {{r_a}} \right)} \right)\left( {{r_{bi}} - E\left( {{r_b}} \right)} \right)} }}{{Std\left( {{r_a}} \right)Std\left( {{r_b}} \right)}}
\end{equation}

At last, as defined in \cite{huang2022relational} - the value of the Spearman's smaller, the stronger negative correlation between the samples. So, to refine the pose estimation accuracy and improve metrics result through minimizing the correlation between MSE loss and pck metrics, the only way is to improve the negative correlation between loss function and PCK metrics: 
\begin{equation}
\label{equation:relation_loss}
\mathcal{{O}}_s\left( {\mathcal{L}\left( {y,\hat y;\theta } \right),\mathcal{M} \left( {y,\hat y} \right)} \right) = {\rho _S}\left( {\mathcal{L}\left( {y,\hat y;\theta } \right), \mathcal{M} \left( {y,\hat y} \right)} \right)
\end{equation}

\noindent where $\theta$ is the optimized parameters in the network. 

\section{Experiments}

\noindent\textbf{Clinical Data Driven Simulation.}
To demonstrate the effectiveness of FJL in stroke rehabilitation, a real clinical rehabilitation dataset, KiMoRe \cite{capecci2019kimore}, was utilized. This dataset encompasses professional rehabilitation exercises for 200 patients across three types of stroke (Upper limb hemiplegia; Parkinson’s syndrome; and back pain syndrome). For simulating robotic rehabilitation motions, the Coppeliasim simulator \cite{rohmer2013v} was employed. A ``robotic stroke rehabilitation" scenario was designed, as depicted in Figure \ref{settings_joints}. Both human and multiple Jaco robotic arm models were adopted. The simulated arm, consisting of 12 joints, guides the rehabilitation movements of the patient's upper limbs. Clinical data from KiMoRe was used to generate human motions within the simulator to best approximate real processes in professional medical rehabilitation. Drawing from these human rehabilitation motions, robot guidance motions were inversely generated using a control-based inverse kinematics (IK) method (introduced in the work \cite{buss2004introduction}), where the robot tracked human forearm position along with the expert's motions. Data from the robot arm, which includes positions, forces, velocities, and orientations of each joint, were meticulously recorded during the simulations. This gathered robot data was subsequently used to train the federated joint learning model. Model accuracy validation can be viewed in Figure \ref{settings_a}. 

\begin{figure}[!t]
	\centering{
	\includegraphics[width=1.0\linewidth]{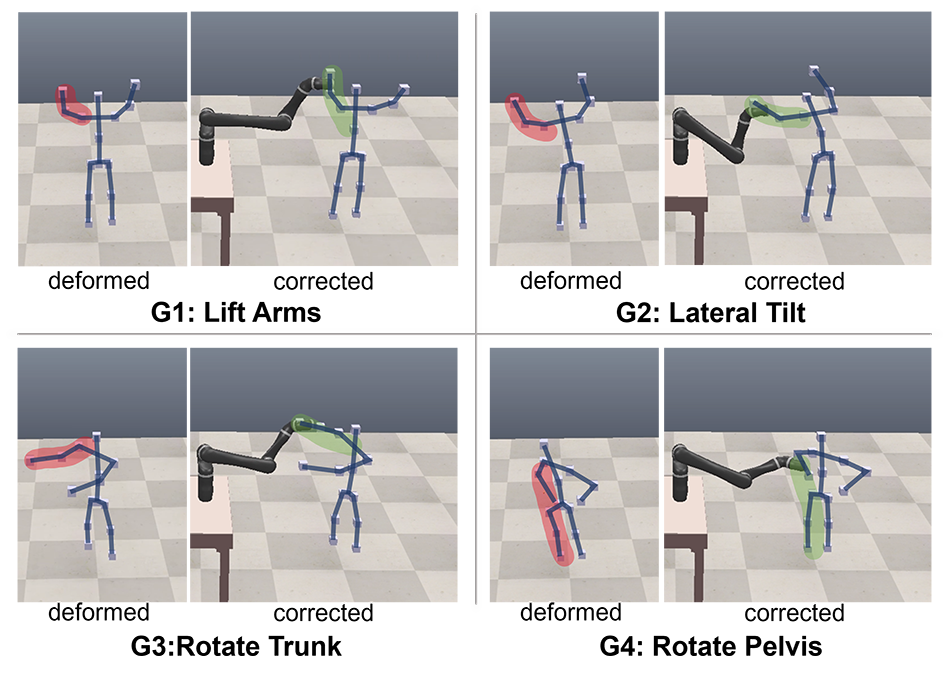}
        \setlength{\belowcaptionskip}{-10pt}
        \setlength{\abovecaptionskip}{-5pt}
	\caption{Four rehabilitation exercises and different action qualities: deformed actions without guidance (red) v.s. corrected actions with guidance (green).}
	\label{taskscenario}
	}
\end{figure}

\noindent\textbf{Task Scenario Design.}
The design of the task in this paper aims to assess whether a robot can aid patients in performing rehabilitation motions accurately, while also adapting to individual patient conditions. We imported expert demonstrations of rehabilitation motions and compared this data with the motions of robot-assisted patients to determine if robotic rehabilitation can achieve results comparable to experts executing the desired motions. To evaluate the efficacy of robot assistance in finer detail, we employed the PCK metrics to assess the robot's pose estimation results in subsequent experiments. We considered four different types of exercises for upper limb rehabilitation support from the dataset: Arm Lifting, Lateral Tilt, Trunk Rotation, and Pelvis Rotations. The task scenario can be seen in Figure \ref{taskscenario}. The FJL Network was trained using a 3080Ti GPU with 12GB VRAM. In total, there are approximately 300,000 data sequences from 200 patients.
   
\subsection{Result Analysis}

\begin{figure}[!t]
\centering
\begin{subfigure}{0.4\textwidth}
    \includegraphics[width=\linewidth]{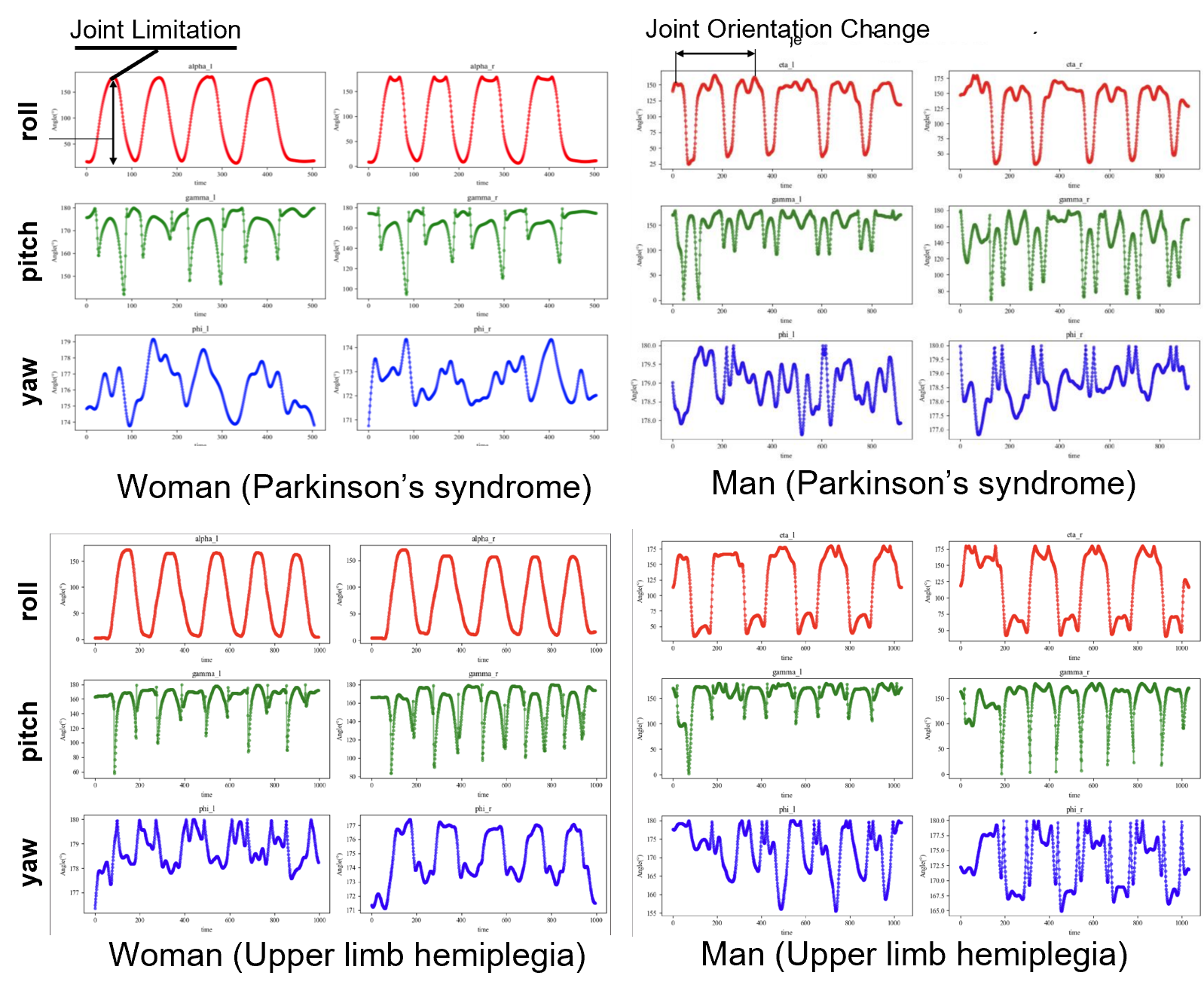}
    \caption{The original patients’ joint pose data. We can infer the human sex, joint limitation and the healthy status from the original data}
\end{subfigure}
\begin{subfigure}{0.4\textwidth}
    \includegraphics[width=\linewidth]{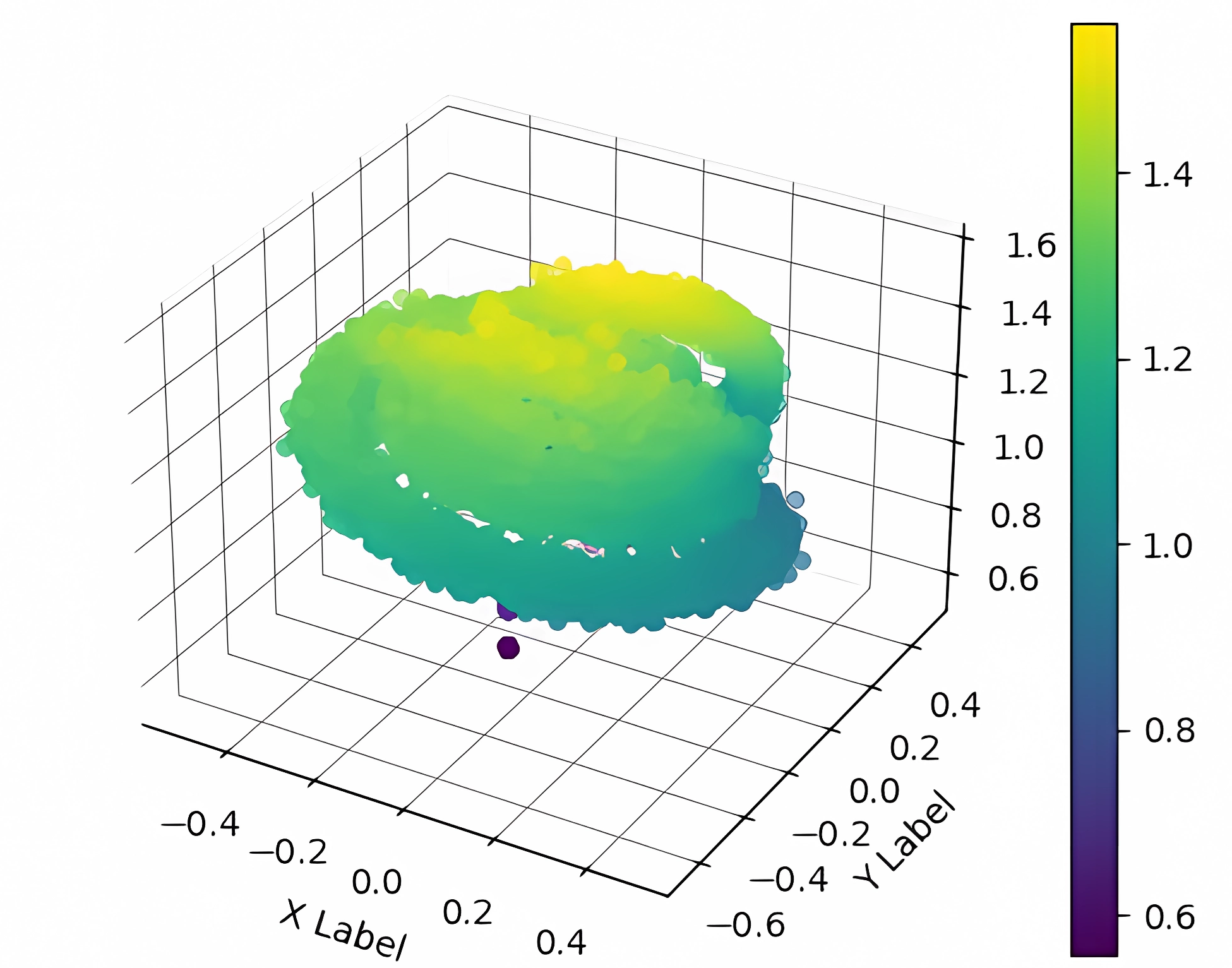}
    \caption{The original patients’ joint pose data distribution. }
\end{subfigure}
\begin{subfigure}{0.4\textwidth}
    \includegraphics[width=\linewidth]{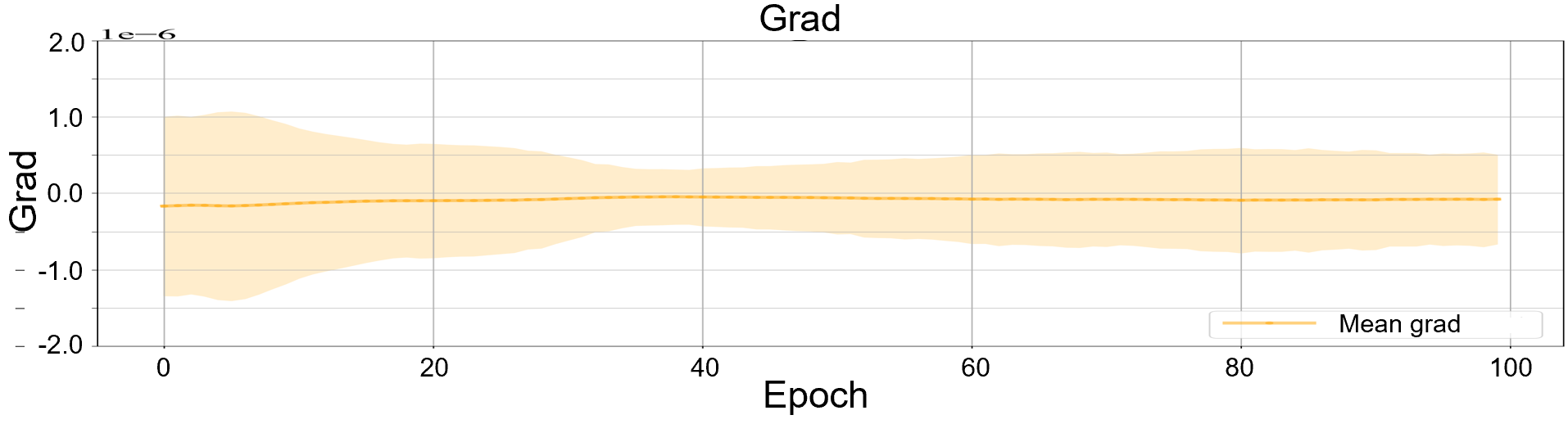}
    \caption{The gradient update without direct information exposure. }
\end{subfigure}
\caption{The transformation from original data to gradient.}
\label{fig:privacy_protection_experiment}

\end{figure}

\textbf{Data Accessibility Analysis.}
The preservation of patient information during the robot training process determines the level of data accessibility. By visualizing and collecting joint data, patient information could be inferred. As shown in Figure \ref{fig:privacy_protection_experiment}(a), patient information regarding gender, age, and stroke types was inferred from the originally-collected joint data, which includes position and roll/pitch/yaw rotations. For instance, a male typically has a rotation angle of 0-153$^\circ$ with an interval period of 200 ms, whereas a female exhibits a rotation angle of 0-160$^\circ$ with an interval period of roughly 100 ms. In \ref{fig:privacy_protection_experiment}(a), we perform the statistical analysis and show the pose coordinates data. In contrast, in the FJL (developed in this study and depicted in \ref{fig:privacy_protection_experiment}(c)), we can see that the 3D data distribution in \ref{fig:privacy_protection_experiment}(b) transforms to the neural network gradient, making it challenging to directly recover and infer patient data. This demonstrates that the FJL facilitates joint learning without directly revealing patient details. Even though the gradients might contain some patient data, the limited information leakage within the FJL framework validates the feasibility of jointly training multiple robots across hospitals.

\begin{figure}[!t]
	\centering{
    \includegraphics[width=1.0\linewidth]{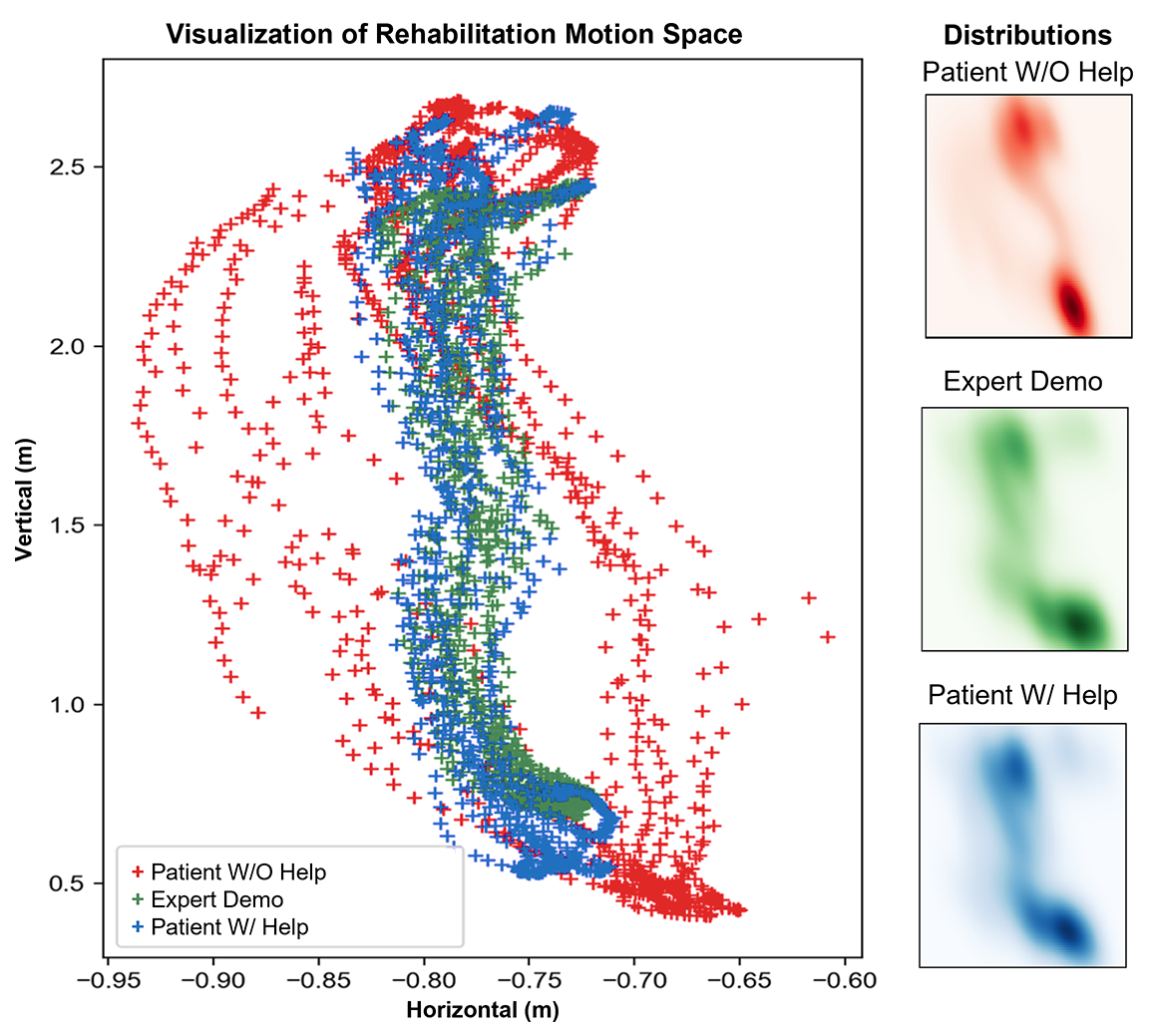}
        \setlength{\belowcaptionskip}{-10pt}
        \setlength{\abovecaptionskip}{0pt}
	\caption{Visualization of rehabilitation motion space in the learned model for three scenarios ``Patient W/O Help", ``Expert Demon", and ``Patient W/ Help"; heat map comparison reveals the action quality difference. 
 }
	\label{fig:robot_assitant_experiment}
	}
\end{figure}

\noindent\textbf{Rehabilitation Model Visualization.} 
The learned rehabilitation skills are visualized in Figure \ref{fig:robot_assitant_experiment}, and the corresponding heatmaps compare the model's performance. Red crosses denote a patient's rehabilitation actions without robot guidance. Blue crosses represent actions with robot assistance. Green crosses depict the standard posture of a healthy human, serving as the ground truth. The heatmaps describe joint distributions learned in rehabilitation models. 

The outcomes of the experiment highlight that with robot assistance, patients can emulate rehabilitation motions similar to those of healthy individuals. Furthermore, the movement limitations of patients who receive robotic assistance show significant improvements compared to those who do not. When aided by the robot assistant, the patients' error is approximately 0.05\textit{m}, below the acceptable threshold of 0.1\textit{m}. In contrast, without robotic aid, the error surges to 0.2\textit{m}, surpassing the acceptable accuracy by a considerable margin. Such imprecise motions not only hinder patients' rehabilitation but can also impede their healing process. The error differential between scenarios with and without robotic assistance is approximately 0.1\textit{m}. The heatmap further reveals that the joint pose coordinate distribution aligns more accurately with the expert demonstration heatmap (ground truth) when aided by the robot. Conversely, without robot assistance, the joint pose coordinates become deformed. These results underline the pivotal role of the rehabilitation robot in enhancing the efficacy of patients' recovery motions and facilitating more effective rehabilitation training.

\begin{figure}[!tbp]
    \centering 

\includegraphics[width=\linewidth]{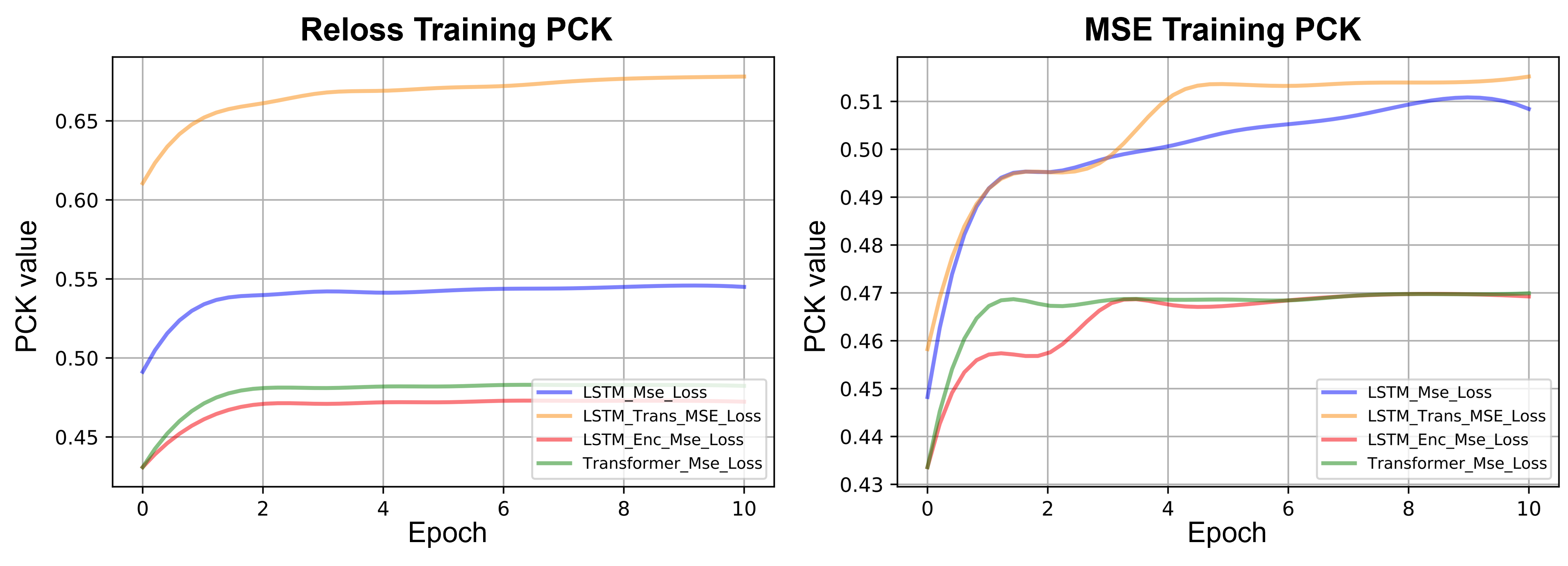}
\setlength{\belowcaptionskip}{5pt}
\setlength{\abovecaptionskip}{-5pt}
\caption{Experiment environment 3D map and trajectory comparison.}
\label{fig:evaluation_scene}
\vspace{-10px}
\end{figure}

\noindent\textbf{Rehabilitation Model Comparison.} To assess the performance of the robot pose estimation module and measure the impact of the Reloss function, we designed and discussed several ablation experiments. As outlined in the FJL Network section, the PCK value was employed to gauge model accuracy. This metric determines whether the Euclidean Distance between the predicted joint pose estimation and the joint pose ground truth is below a specified threshold. As detailed in the Reloss function subsection, this threshold is set at 0.1\textit{m}. A higher PCK value indicates superior performance. Our evaluation comprised two main steps: first, comparing the training curvature for each PCK value and second, conducting the Reloss ablation experiment. The results of these experiments are presented in Figure \ref{fig:evaluation_scene} and Table \ref{tab:evaluation_test_set}.

From Figure \ref{fig:evaluation_scene}, the initial experiment contrasts the efficacy of different modules in training the robot pose estimation network. Four distinct models are compared: the single LSTM, the single Transformer, the LSTM-Encoder-Decoder, and the LSTM-Transformer. When these algorithms utilize the MSE loss to train the network, the PCK metric value of the proposed LSTM-Transformer surpasses other baselines by 20\%-30\%. 

In the relational loss ablation test, it was evident that implementing Reloss in an individual model, rather than the MSE loss, resulted in a performance uptick ranging between 10\% and 15\% when juxtaposed with other control groups. Furthermore, this usage promotes a steadier training curve for the metrics during the training phase. This data reinforces the benefits of recognizing the inverse correlation between MSE loss and metrics, as it not only amplifies overall training performance but also ensures a consistent training metrics curve, ultimately yielding enhanced training results.

The parameters exhibiting the most optimal performance were preserved throughout the training regimen. These parameters were subsequently deployed on the test set for evaluation. The conclusive outcomes are delineated in Table \ref{tab:evaluation_test_set}.

\begin{table}[tbp]
\centering
\caption{The comparison of PCK value of each model with relational loss and MSE loss}
\resizebox{\linewidth}{!}{
\begin{tabular}{cccccccc}
\toprule
 \shortstack{VIO Algorithms} &\shortstack{MSE Loss} & \shortstack{\textbf{Relational Loss (Ours)}}\\
\midrule
LSTM & 0.674 & \textbf{0.751}\\
\midrule

Transformer & 0.565 & \textbf{0.630}\\
\midrule

LSTM-Encoder-Decoder & 0.541 & \textbf{0.743} \\
\midrule

LSTM-Transformer & 0.706 & \textbf{0.754} \\
\midrule

\end{tabular} 
}

\label{tab:evaluation_test_set}
\end{table}

From the results presented in Table \ref{tab:evaluation_test_set}, the superior accuracy and impressive generalization capabilities of the LSTM-Transformer network proved good performance in robot pose estimation. The experiment result indicates the performance of LSTM-Transformer is 20\%-30\% higher than the baseline.

\section{CONCLUSIONS}

This research tackled the nuanced challenges associated with federated joint learning in the context of rehabilitation robot training. We introduced the FJT, a federated learning architecture specifically tailored to foster joint learning amongst networked robots. Notably, this design enables robotic rehabilitation training without directly accessing patient data. Moreover, it paves the way for a collaborative training framework for a robot network across multiple medical facilities without jeopardizing patient confidentiality. Our approach is underpinned by a clinical-data-driven simulation tailored to rehabilitating patients suffering from upper limb hemiplegia due to strokes. The efficacy of FJT has been substantiated in several areas, namely, safeguarding patient-sensitive data, enabling effective robot joint training, and fostering efficient spatial-temporal rehabilitation action learning. As we look ahead, our focus will shift toward the formulation of multi-agent collaborative mechanisms that facilitate the synchronized rehabilitation of both the upper and lower limbs of affected patients.
  



\bibliographystyle{IEEEtran}
\bibliography{reference}
\end{document}